\documentclass{article}

\usepackage[accepted]{mlsys2025}

\usepackage{amsmath,amssymb,mathtools}
\usepackage{booktabs}
\usepackage{siunitx}
\usepackage{graphicx}
\usepackage{subcaption}

\usepackage{adjustbox}
\newcommand{\fitcol}[1]{\begin{adjustbox}{max width=\linewidth}#1\end{adjustbox}}

\usepackage{tikz}
\usetikzlibrary{positioning,arrows.meta,calc,fit,shapes.misc,shapes.geometric}
\usepackage{pgfplots}
\pgfplotsset{compat=1.18}
\usepgfplotslibrary{groupplots}

\usepackage{xcolor}
\usepackage{enumitem}
\usepackage{microtype}
\usepackage{xspace}
\usepackage{comment}
\usepackage{url}

\usepackage{algorithm}                
\makeatletter
\@ifpackageloaded{algorithmic}{%

}{}%
\makeatother
\usepackage[noend]{algpseudocode}     
\algrenewcommand\algorithmicrequire{\textbf{Input:}}
\algrenewcommand\algorithmicensure{\textbf{Output:}}

\usepackage[hidelinks]{hyperref}
\definecolor{inputcol}{RGB}{230,242,255}
\definecolor{dwcol}{RGB}{217,235,252}
\definecolor{pwkeep}{RGB}{214,245,214}
\definecolor{pwsynth}{RGB}{255,236,204}
\definecolor{headcol}{RGB}{233,214,252}
\definecolor{gencol}{RGB}{255,214,214}
\definecolor{cachecol}{RGB}{230,230,230}
\definecolor{traincol}{RGB}{230,242,255}
\definecolor{emacol}{RGB}{255,236,204}
\definecolor{calibcol}{RGB}{232,255,232}
\definecolor{selectcol}{RGB}{242,229,255}
\definecolor{deploycol}{RGB}{255,228,225}

\tikzset{
  cblk/.style={
    draw,
    rounded corners=2pt,
    inner sep=3pt,
    minimum height=6.5mm,
    align=center,
    font=\scriptsize
  }
}

\newcommand{\method}{\textsc{HyperTinyPW}\xspace}

\newcommand{\dw}{DW\xspace}
\newcommand{\pw}{PW\xspace}
\newcommand{\vae}{VAE\xspace}
\newcommand{\kb}{\,kB\xspace}
\newcommand{\mb}{\,MB\xspace}

\title{Once-for-All Channel Mixers (\textsc{HyperTinyPW}): Generative Compression for TinyML}
\date{}
\author{
  Yassien Shaalan \\
  \texttt{yassien@gmail.com}
}
\begin{document}
\maketitle
\begin{abstract}
Deploying neural networks on microcontrollers is constrained by kilobytes of flash/SRAM, where $1{\times}1$ pointwise (PW) mixers often dominate memory even after \texttt{INT8} quantization across vision, audio, and wearable sensing. We present \method{}, a \emph{compression-as-generation} approach that replaces most \emph{stored} PW weights with \emph{generated} weights: a shared micro-MLP synthesizes PW kernels once at load time from tiny per-layer codes, caches them, and executes them with standard integer operators. This preserves commodity MCU runtimes and adds only a one-off synthesis cost; steady-state latency/energy match \texttt{INT8} separable CNN baselines. Enforcing a shared latent basis across layers removes cross-layer redundancy, while keeping PW$_1$ in \texttt{INT8} stabilizes early, morphology-sensitive mixing. We contribute (i) TinyML-faithful \emph{packed-byte} accounting covering generator, heads/factorization, codes, kept PW$_1$, and backbone; (ii) a unified evaluation with validation-tuned $t^\star$ and bootstrap CIs; and (iii) a deployability analysis (integer-only inference, boot vs.\ lazy synthesis).
On three ECG benchmarks (Apnea-ECG, PTB-XL, MIT-BIH), \method{} shifts the macro-$F_1$–vs.–flash Pareto: at $\sim$225\,kB it matches a $\sim$1.4\,MB CNN while being $6.31{\times}$ smaller (84.15\% fewer bytes), retaining $\geq\!95\%$ of large-model macro-$F_1$. Under 32–64\,kB budgets it sustains balanced detection where compact baselines degrade. The mechanism applies broadly to other 1D biosignals, on-device speech, and embedded sensing tasks where per-layer redundancy dominates, indicating a wider role for \emph{compression-as-generation} in resource-constrained ML systems.
\end{abstract}

\section{Introduction}

Deep learning models for biosignal analytics, such as ECG, are increasingly expected to run \emph{directly on microcontrollers (MCUs)}. On-device inference improves \emph{privacy, reliability, and energy proportionality}, since data never leaves the sensor and decisions can be made in real time. Yet MCUs offer only tens of kilobytes of flash and SRAM and limited compute extensions (e.g., Arm M-series DSP). These constraints create a sharp tension between the promise of local analytics and the cost of deploying modern convolutional neural networks.

Among TinyML backbones, \emph{separable 1D CNNs} are attractive: depthwise (DW) convolutions dominate multiply–accumulates, while \emph{pointwise (PW, 1$\times$1) convolutions} concentrate most parameters. Unfortunately, even after \texttt{INT8} quantization, multiple PW layers still dominate flash usage, often pushing total storage beyond 64\,kB. Thus, the PW mixers-not the depthwise layers-become the limiting factor for MCU deployment.

Classical compression techniques-quantization, pruning, low-rank or tensor factorization-shrink parameters but still \emph{store a full set of weights for every PW layer}. Dynamic weight generation (HyperNetworks, DFN, CondConv) can reduce redundancy but typically \emph{generate kernels per input}, adding branching, SRAM pressure, and latency jitter incompatible with real-time MCU workloads. What is missing is a strategy that \emph{directly targets the PW bottleneck} while respecting strict device constraints: no per-example branching, minimal SRAM, and unmodified integer kernels.

We propose to replace \emph{stored PW weights} with \emph{generated weights once per layer at load time}. Our method, \method{}, uses a shared micro-MLP to synthesize most PW kernels from tiny per-layer codes, while deliberately keeping the first PW (PW$_1$) in \texttt{INT8} to anchor morphology-sensitive early mixing. Generation occurs once-at boot or lazily when first needed-then synthesized weights are cached and reused. Inference proceeds entirely with \emph{standard integer operators}, ensuring compatibility with existing TinyML stacks. We also contribute a \emph{TinyML-faithful packed-byte accounting} that includes the generator, its heads, the retained PW$_1$, codes, and the backbone.

Beyond storage savings, the design enforces a \emph{shared latent basis across layers}. This reduces redundancy and acts as an \emph{efficient-coding prior}: layers reuse common generative factors rather than relearning separate mixers. From a representation-learning perspective, this behaves like \emph{implicit multi-task regularization across layers}, helping preserve balanced detection even under tight flash budgets. An analogy is skill reuse in humans-experts deploy compact routines (rhythmic or grammatical motifs) across many tasks. Similarly, our generator emits shared transformations that multiple layers can repurpose.

Beyond compression, our goal is to make channel mixing \emph{deployable} on real MCUs. We therefore (i) report \emph{packed-byte} sizes that match what ships; (ii) keep inference on unmodified integer kernels (CMSIS-NN/TFLM compatible); and (iii) characterize \emph{boot vs.\ lazy} synthesis, peak SRAM, and the latency/energy impact of one-shot generation. This positions cross-layer generative mixing as a \emph{systems design} for TinyML, not just a model variant.

We validate the approach on three representative ECG tasks-Apnea-ECG (minute-level apnea detection), PTB-XL (diagnostic proxy), and MIT-BIH (AAMI arrhythmia grouping)-using \emph{record/patient-wise splits} to avoid identity leakage. Experiments include comprehensive ablations (hybrid vs.\ all-synth, latent sizes $(d_z,d_h,r)$, precision 4--8\,bit, knowledge distillation, focal loss), structured alternatives under equal flash, and system-level boot-time and SRAM measurements. Finally, we analyze the \emph{accuracy--memory Pareto frontier}, offering guidance for MCU-constrained deployment.

\textbf{Contributions.}
(1) \textbf{Compression-as-generation for PW layers.} A shared micro-MLP synthesizes most $1{\times}1$ mixers from tiny per-layer codes once at load time, while PW$_1$ remains \texttt{INT8} to stabilize early morphology; steady-state inference uses standard integer kernels (no custom ops).
(2) \textbf{TinyML-faithful accounting and deployment.} Exact \emph{packed-byte} sizes (generator, heads/factorization, codes, kept PW$_1$, backbone), plus a deployment analysis of \emph{boot} vs.\ \emph{lazy} synthesis, SRAM peaks, and compatibility with CMSIS-NN/TFLM.
(3) \textbf{Latency/energy profiling.} A lightweight pipeline that reports per-inference latency and energy on MCU backends (virtual and on-device), isolating generation overhead from steady-state compute.
(4) \textbf{Rigorous validation under MCU budgets.} Cross-dataset results (Apnea-ECG, PTB-XL, MIT-BIH), ablations over $(d_z,d_h,r)$ and precision (4–8\,bit), and Pareto curves of macro-$F_1$ vs.\ packed flash that expose the ${\sim}225$\,kB elbow. 

At ${\sim}225$\,kB, \method{} compresses a $1.4$\,MB baseline by $6.31{\times}$ (84.15\% fewer bytes) while retaining ~$95\%$ of its macro-$F_1$ on Apnea-ECG and PTB-XL. This places \method{} at the mid-budget elbow of the accuracy–flash Pareto while preserving a TinyML-faithful deployment path (packed-byte accounting, one-shot load-time synthesis, and integer-only inference).

\section{Related Work}
\label{sec:related}

Our work connects four areas: TinyML backbones, compression of channel mixing, dynamic weight generation, and ECG deep learning. Below we situate \method{} in each thread and address likely counter-arguments.

\textbf{Tiny models and MCU deployment.}
MobileNet-style backbones \citep{howard2017mobilenets,sandler2018mobilenetv2,howard2019mobilenetv3} reduce MACs via depthwise (\dw{}) layers but concentrate parameters in $1{\times}1$ pointwise (\pw{}) mixers. MCU software stacks (CMSIS-NN, TFLM) enable \texttt{INT8} inference \citep{lai2018cmsisnn}, MLPerf Tiny standardizes tasks \citep{banbury2021mlperf}, and co-design (MCUNet, Once-for-All) explores NAS for fit/latency \citep{lin2020mcunet,cai2020onceforall}. One might argue that NAS can simply remove or shrink \pw{} layers. Empirically, completely eliminating \pw{} mixing collapses channel reuse and hurts accuracy; even after NAS, remaining \pw{} layers dominate flash. \method{} is orthogonal: it retains \pw{} expressivity but amortizes storage across layers.

\textbf{Compression methods.}
Quantization \citep{jacob2018quantization}, pruning/sparsity \citep{han2016deepcompression,frankle2019lottery}, and low-rank/tensor factorization \citep{denton2014exploiting} reduce parameters, yet the model still \emph{stores} a parameterization per \pw{} layer. In the 32–64\,KB regime, several \pw{} matrices remain the bottleneck. A counter-claim is that aggressive low-rank or $k$-sparse \pw{} will suffice. However, these approaches attack \emph{within-layer} redundancy and do not capture \emph{cross-layer} regularities; they also incur per-layer metadata that adds up at TinyML scale. \method{} instead ties layers through a shared generator and tiny layer codes, and can \emph{compose} with low-rank/sparse heads (we evaluate structured baselines at equal flash).

\textbf{Structured transforms.}
Algebraic operators (circulant, Toeplitz, Kronecker, ACDC) \citep{sindhwani2015structured,moczulski2016acdc} shrink matrices and sometimes speed up inference. It may seem these remove the need for generation. In practice, they restrict each \pw{} \emph{independently}, leaving the per-layer storage pattern intact; they also require kernel support to realize speedups on MCUs. Our approach is complementary: structured heads can further compress $H_l$ (or $A_l,B$), but the novelty is \emph{cross-layer} synthesis that leaves inference kernels unchanged.

\textbf{Dynamic and generated weights.}
HyperNetworks \citep{ha2017hypernetworks}, dynamic filter networks \citep{jia2016dfn}, CondConv \citep{yang2019condconv}, and dynamic convolution \citep{chen2020dynamicconv} generate or mix kernels \emph{per input}, which suits GPUs/TPUs. A natural question is whether small dynamic modules could also fit MCUs. The barrier is not only parameter count but per-input control flow, SRAM peaks, and latency jitter, which violate real-time/energy budgets. \method{} generates \emph{once at load time}, caches weights, and then runs standard \texttt{INT8} inference with no runtime control changes.

\textbf{ECG deep learning and TinyML.}
Large ECG models achieve strong accuracy \citep{rajpurkar2017cardiologist,hannun2019nature,ribeiro2020ncomm} on datasets such as PTB-XL \citep{wagner2020ptbxl} and MIT-BIH \citep{moody2001mitbih}, but embedded deployments often omit \emph{packed-byte} accounting or exceed MCU flash. One might claim that downsampling or binarization makes any model fit. In practice, these shortcuts degrade morphology-sensitive detection and still leave multiple \pw{} layers too large. We explicitly report deployable packed bytes (generator, heads, codes, kept PW$_1$, backbone) and demonstrate record/patient-wise generalization within 32–64\,KB.

\textbf{Positioning.}
Relative to compression, \method{} avoids per-layer storage by synthesizing most \pw{} weights from a shared micro-MLP and tiny codes; relative to structured transforms, it introduces cross-layer parameter tying that can coexist with algebraic heads; relative to dynamic methods, it eliminates per-input cost by performing \emph{layer-constant} generation at load time; relative to prior ECG TinyML, it provides TinyML-faithful accounting and a deployable path under strict MCU budgets. The combination—shared generator, keep-first-\pw{} hybrid, and packed-byte evaluation—appears novel in MCU-scale biosignal inference.

\section{Method}
\label{sec:method}

Our goal is to compress separable CNNs for microcontrollers by replacing most stored
pointwise (\pw{}) channel mixers with compact, generated representations. Instead of
storing every \pw{} kernel, we store only a tiny per-layer code and use a shared generator
to expand these codes into full kernels once at load time. This preserves compatibility
with standard \texttt{INT8} inference while tying layers together through a shared latent
basis. 

The section is organized as follows. We first introduce the setup and notation for separable
CNNs. We then describe our generative channel mixing approach, including how codes,
generator, and heads interact. Next, we detail deployable storage via packed-byte accounting,
present the training objective, and explain the calibration pipeline. Finally, we outline MCU
deployment options, show the load-time synthesis algorithm, and compare complexity to
conventional \pw{} stacks.

\subsection{Setup and notation}
We consider a compact separable 1D CNN for ECG: each block applies a depthwise temporal convolution (\dw{}) followed by a $1{\times}1$ \pw{} channel mixer. Let $x \in \mathbb{R}^{C_{\mathrm{in}}\times T}$ be the input (channels $\times$ time). A \pw{} layer $l$ multiplies the channel dimension by
\(
W_l \in \mathbb{R}^{C_{\mathrm{out}}^{(l)} \times C_{\mathrm{in}}^{(l)}}
\)
(bias omitted). In conventional TinyML deployment, every $W_l$ is stored (typically \texttt{INT8}), and the sum $\sum_l C_{\mathrm{out}}^{(l)}C_{\mathrm{in}}^{(l)}$ dominates flash.

\subsection{Generative channel mixing (layer-constant synthesis)}
Instead of storing each full \pw{} matrix, we assign a small \emph{code} $z_l$ to each layer and let a shared generator produce its weights once at load time. Crucially, generation is \emph{layer-constant}, never per input: the synthesized weights are cached and then reused by standard integer kernels.

Each code $z_l \in \mathbb{R}^{d_z}$ is mapped by the generator $g_\phi$ into an embedding:
\begin{equation}
\label{eq:gen-embed}
h_l = g_\phi(z_l),
\end{equation}
where $h_l \in \mathbb{R}^{d_h}$ is a compact hidden vector summarizing layer $l$.  

A light per-layer head $H_l$ then projects this embedding into the vectorized kernel and reshapes it into the full PW matrix:
\begin{equation}
\label{eq:head-project}
\widehat{w}_l = H_l h_l \in \mathbb{R}^{C_{\mathrm{out}}^{(l)} C_{\mathrm{in}}^{(l)}}, 
\qquad
\widehat{W}_l = \mathrm{reshape}\!\big(\widehat{w}_l,\; C_{\mathrm{out}}^{(l)}, C_{\mathrm{in}}^{(l)}\big).
\end{equation}

To compress further, $H_l$ can be factorized into a small per-layer adapter $A_l$ and a shared matrix $B$:
\begin{equation}
\label{eq:head-factor}
H_l = A_l B, \quad
A_l \in \mathbb{R}^{(C_{\mathrm{out}}^{(l)} C_{\mathrm{in}}^{(l)}) \times r}, \;
B \in \mathbb{R}^{r \times d_h}, \; r \ll d_h.
\end{equation}
This means most of the capacity lives in $B$ (shared), while each layer only stores its lightweight adapter $A_l$.  

Because early mixing is morphology-sensitive, we deliberately keep PW$_1$ as a stored \texttt{INT8} layer and synthesize only PW$_{2:L}$.

\begin{figure*}[t]
  \centering
  \includegraphics[
    width=\textwidth,
    height=0.82\textheight,
    keepaspectratio
  ]{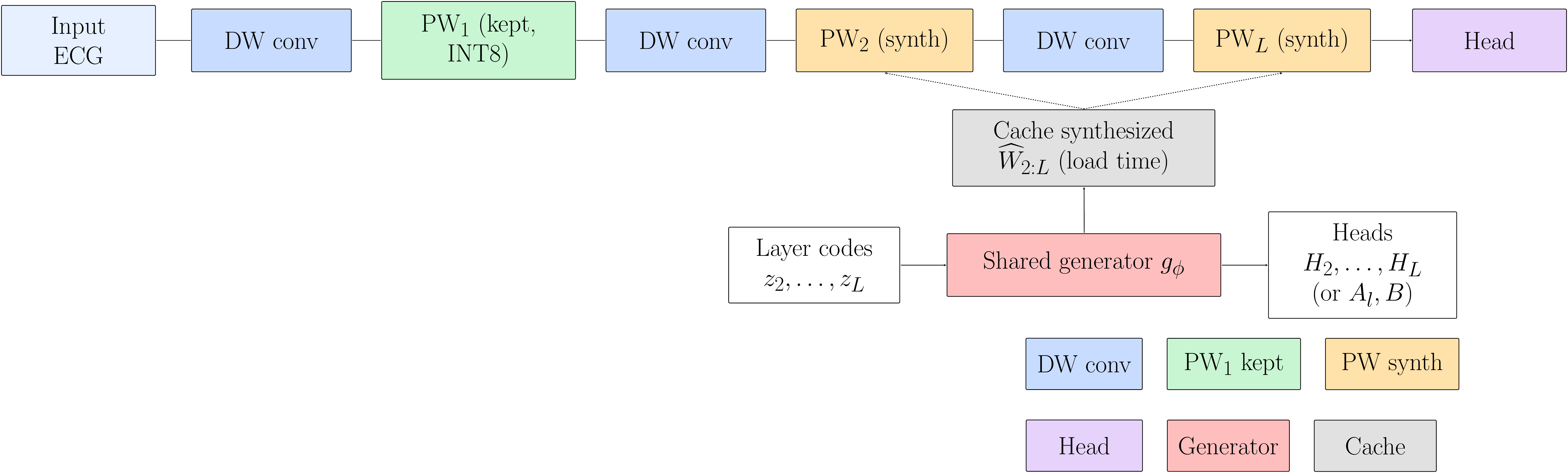}
  \vspace{-2mm}
\caption{\textbf{Architecture overview.} Depthwise (blue), PW$_1$ kept in INT8 (green), synthesized PW layers (orange), classifier head (purple). The shared generator (red) produces PW$_{2:L}$ once at load time and caches them (gray); steady-state inference uses standard integer kernels.}
  \label{fig:arch-colored}
\end{figure*}

\begin{figure*}[t]
  \centering
  \includegraphics[width=0.8\textwidth]{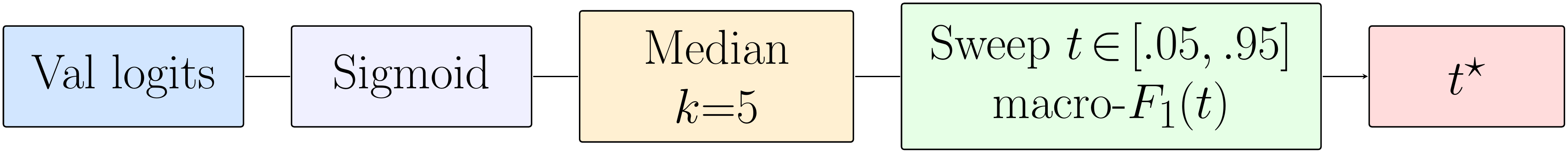}
  \caption{Validation thresholding pipeline.}
  \label{fig:calibration}
\end{figure*}

\subsection{Deployable storage: packed-byte accounting}
We carefully count the deployable flash footprint. For each tensor $\tau$ with $N_\tau$ elements stored at bitwidth $b_\tau$, the footprint is
\begin{equation}
\label{eq:flash}
\left\lceil \frac{N_\tau\, b_\tau}{8} \right\rceil \;\;\text{bytes}.
\end{equation}
The total flash is the sum across generator parameters, heads (or factorized $A_l,B$), codes, the kept PW$_1$, and backbone layers.  
We quantize $\{\phi, H_l, z_l\}$ to 4/6/8\,bits, while keeping stem, DW, PW$_1$, and classifier at \texttt{INT8}. This ensures a fair, deployable accounting.

\subsection{End-to-end training (stability, size, and imbalance-aware)}
We train the generator and student jointly with AdamW, applying GN(1) instead of BN for small-batch stability, NaN-safe initialization, gradient clipping, and an EMA of weights. The composite loss is
\begin{equation} \begin{aligned} \mathcal{L} &= \mathrm{CE}(y,\hat{y})\\ &\quad + \lambda_{\mathrm{foc}}\,\mathrm{Focal}_\gamma(y,\hat{y})\\ &\quad + \lambda_{\mathrm{KD}}\,\mathrm{KL}\bigl(\sigma(\hat{y}/T)\,\|\,\sigma(\hat{y}^{\text{teach}}/T)\bigr)\\ &\quad + \lambda_{\mathrm{feat}}\,\bigl\lVert \hat{f}-\hat{f}^{\text{teach}} \bigr\rVert_2^2\\ &\quad + \lambda_{\mathrm{softF1}}\,\mathcal{L}_{\mathrm{softF1}}(\hat{y},y)\\ &\quad + \lambda_{\mathrm{spec}}\,\mathcal{R}_{\mathrm{spec}}(\theta)\\ &\quad + \lambda_{\mathrm{size}}\,\bigl\lVert \theta_{\text{heads,codes}} \bigr\rVert_1. \end{aligned} \label{eq:task-loss} \end{equation}
where CE drives baseline accuracy, focal loss counteracts class imbalance by emphasizing hard or minority examples, KL distillation and feature matching transfer knowledge from a larger teacher, soft-$F_1$ optimizes directly for the evaluation metric, spectral regularization stabilizes dynamics by constraining layer smoothness, and the $L_1$ penalty enforces compact codes and heads to reduce flash usage; jointly, these terms promote stability, imbalance-awareness, and size efficiency, enabling \method{} to converge reliably while remaining sensitive to rare events under microcontroller constraints. Unlike prior TinyML training objectives that rely on only CE or CE+KD, this formulation unifies metric-aware, imbalance-aware, and compression-aware terms into a single end-to-end objective, making the loss itself a vehicle for co-designing accuracy, robustness, and deployability.

\subsection{MCU deployment: boot vs.\ lazy synthesis}
Synthesis runs once per layer and cached weights are reused thereafter.  
Two schedules are possible: \emph{boot synthesis}, generating all PW$_{2:L}$ at startup (faster inference, longer boot), or \emph{lazy synthesis}, generating each PW$_l$ on first use (shorter boot, one-time stall). 
Steady-state inference always uses \texttt{INT8} kernels; $g_\phi$ is never called per input. 
Peak SRAM is bounded by the largest PW plus activations, and weights can be streamed to flash if needed.
Regarding Kernel compatibility, synthesis emits \texttt{INT8}-quantized PW tensors laid out to match standard $1{\times}1$ conv GEMV paths in CMSIS-NN/TFLM. No graph rewrites or per-example control flow are introduced. As a result, deployment reduces to a one-time “weight install” followed by vanilla integer inference.

\begin{algorithm}[t]
\caption{Load-time synthesis and caching (MCU side)}
\label{alg:synth}
\begin{algorithmic}[1]
\For{$l = 2$ \textbf{to} $L$}
  \State $h_l \gets g_\phi(z_l)$ \hfill\textit{(Eq.~\ref{eq:gen-embed})}
  \State $\widehat{w}_l \gets H_l h_l$ \hfill\textit{(or $A_l B h_l$, Eq.~\ref{eq:head-project}--\ref{eq:head-factor})}
  \State $\text{PW}_l \gets \mathrm{reshape}(\widehat{w}_l)$; cache $\text{PW}_l$
\EndFor
\State \textbf{Inference:} run \texttt{INT8} DW/PW with cached $\{\text{PW}_l\}$; no calls to $g_\phi$ per input.
\end{algorithmic}
\end{algorithm}

\subsection{Complexity and storage}
A standard \pw{} stack stores $\sum_l C_{\mathrm{out}}^{(l)} C_{\mathrm{in}}^{(l)}$ \texttt{INT8} parameters.  
In \method{}, the stored footprint becomes
\begin{multline}
\mathrm{flash\_total}
= |\phi|
+ \Big(\sum_l |H_l|\ \text{or}\ \sum_l |A_l| + |B|\Big)
+ \sum_l |z_l| \\
+ |PW1|
+ |{\rm stem,DW,head}|.
\end{multline}

With $(d_z,d_h,r) \ll C_{\mathrm{out}}^{(l)} C_{\mathrm{in}}^{(l)}$, the packed bytes (Eq.~\ref{eq:flash}) drop sharply while inference cost matches the baseline. Boot/lazy synthesis adds a one-off cost proportional to the total PW size.

\subsection{VAE-head baseline (decoder-free at deploy)}
As an additional baseline, we evaluate a lightweight 1D \vae{} encoder whose latent $z$ is classified by a small MLP. 
The decoder is used only during training for latent regularization and reconstruction consistency, then discarded at deployment; 
the stored footprint therefore includes only the encoder and classification head, reported under the same packed-byte accounting. 
This baseline was originally considered to complement our generative synthesis approach by testing whether a learned latent manifold could implicitly encode meaningful structural priors without explicit parameter generation. 
However, while the VAE-head achieves competitive compactness, it lacks the ability to synthesize or adapt weights across layers, and its latent space regularization often trades off discriminative sharpness for reconstruction fidelity. 
As a result, it serves mainly as a contrastive probe showing that generative \emph{weight synthesis}, rather than purely latent compression, is responsible for the gains observed in \method{}.

\section{System Profiling and Deployment}
\label{sec:system}

\textbf{Targets and kernels.}
We target Arm M-series MCUs running TFLM with CMSIS-NN integer kernels; all PW layers execute through stock $1{\times}1$ conv/GEMV paths. Our method requires no custom ops: synthesized PW tensors are cached as ordinary INT8 weights.

\textbf{Profiling methodology (proxies).}
We separate one-shot load-time synthesis from steady state. Steady-state latency is obtained from an instruction/cycle model of the CMSIS-NN/TFLM kernels on a virtual MCU configuration (e.g., Arm FVP / Renode / QEMU). Energy per inference is derived from cycles via a calibrated board-level current model (normalized nJ/inference). These are \emph{hardware-agnostic proxies} intended for model-to-model comparison rather than absolute device benchmarking.

\textbf{Boot vs.\ lazy synthesis.}
\emph{Boot} compiles PW$_{2:L}$ at startup (higher startup time, no first-inference stall). \emph{Lazy} compiles on first use (fast boot, one-time per-layer stall). Both schedules yield identical steady-state latency because inference runs entirely on cached INT8 weights.

\textbf{SRAM peaks and streaming.}
Peak SRAM is the maximum of \{largest PW activation tensor, workspace\}. If writable flash is available, synthesized PW tensors can be streamed back to flash to cap SRAM peaks; otherwise we synthesize layer-wise with buffer reuse.

\textbf{Portability.}
Because inference uses unmodified integer kernels and generation happens only at load time, the approach ports to any TFLM-compatible stack with INT8 convolutions. We release scripts to regenerate packed-byte counts and cycle/energy proxies.

\section{Experimental Setup}
\label{sec:setup}

\subsection{Datasets \& Preprocessing}
\label{sec:data}
We choose three \emph{single-lead} ECG corpora that jointly cover screening, clinical diagnostics, and arrhythmia detection while stressing MCU constraints (short windows, low-rate inputs, and class priors from balanced to highly skewed).

\textbf{Apnea-ECG} (PhysioNet~\cite{ichimaru1999apnea, goldberger2000physionet}). 
Originally collected for sleep apnea screening, this dataset provides single-lead overnight ECGs with minute-wise apnea annotations. 
We segment each record into 18\,s windows at 100\,Hz, apply per-window $z$-score normalization with a variance guard, 
and split 80/10/10 by \emph{record/patient} to avoid identity leakage. 
Its label distribution is notably skewed toward non-apnea segments, making it a good testbed for imbalance-aware training.  

\textbf{PTB-XL}~\citep{wagner2020ptbxl}. 
A large, heterogeneous clinical collection (21{,}837 records from 18{,}885 patients) with diverse pathologies, 
recorded at both 500 and 100\,Hz. 
We downsample to 100\,Hz, use lead~II, binarize labels (NORM vs.\ any diagnostic superclass), 
extract 10\,s windows, and re-materialize 8/1/1 folds for train/val/test. 
Its scale, label heterogeneity, and real-world noise make it a strong benchmark for generalization under compression.  

\textbf{MIT-BIH Arrhythmia} (PhysioNet~\cite{moody2001mitbih,goldberger2000physionet}). 
A long-standing reference corpus of arrhythmia recordings with beat-level annotations from 47 subjects. 
We adopt the AAMI binary setup (normal vs.\ arrhythmia) with single-lead inputs, following common tiny-ECG practice~\cite{aami1998ec57}. 
This dataset is compact but challenging, with highly imbalanced arrhythmia types and patient-specific morphology differences.  

All datasets share a consistent windowing/normalization policy and retain source sampling rates, 
allowing fair cross-dataset comparisons under identical TinyML constraints.

\begin{table}[t]
\centering
\small
\caption{Class priors (\%) per split (Class 0 = negative/normal, Class 1 = positive/event). These priors inform validation-tuned thresholding and help interpret AUC vs.\ macro-$F_1$ behavior.}
\label{tab:class-balance}
\begin{tabular}{lccc}
\toprule
Dataset (Split) & Class 0 (\%) & Class 1 (\%) & Samples \\
\midrule
\textbf{Apnea-ECG} (Train) & 50.25 & 49.75 & 2000 \\
\textbf{Apnea-ECG} (Val)   & 33.85 & 66.15 & 2000 \\
\textbf{Apnea-ECG} (Test)  & 38.40 & 61.60 & 2000 \\
\midrule
\textbf{PTB-XL} (Train) & 53.20 & 46.80 & 2000 \\
\textbf{PTB-XL} (Val)   & 44.35 & 55.65 & 2000 \\
\textbf{PTB-XL} (Test)  & 43.80 & 56.20 & 2000 \\
\midrule
\textbf{MIT-BIH} (Train) & 92.95 & 7.05 & 2000 \\
\textbf{MIT-BIH} (Val)   & 90.20 & 9.80 & 2000 \\
\textbf{MIT-BIH} (Test)  & 93.50 & 6.50 & 2000 \\
\bottomrule
\end{tabular}
\end{table}

This mix provides complementary regimes-balanced (Apnea-ECG train/test), mildly skewed (PTB-XL), and highly imbalanced (MIT-BIH)—so we can study accuracy–flash trade-offs and calibration under deployment-faithful conditions. The strong skew in MIT-BIH (Table~\ref{tab:class-balance}) motivates our use of macro-$F_1$, balanced accuracy,

\subsection{Model Suite \& Sweep (21 runs per dataset)}
\label{sec:models}
\textbf{\method{} (ours).} We replace most PW mixers with weights generated once per layer from tiny codes, keeping PW$_1$ in \texttt{INT8}. We sweep $(d_z,d_h)\!\in\!\{(4,12),(6,16)\}$, bit-widths $\{8,6\}$ for $\{\phi,H,z\}$, and (optionally) KD $\Rightarrow$ 8 runs.
\textbf{TinyVAE-Head.} A lightweight DW/PW encoder with a VAE head ($q_\psi(z|h)$, reparameterized $z$) feeding a tiny classifier; focal on; KD on/off; $\{8,6\}$-bit $\Rightarrow$ 4 runs.
\textbf{CNN/ResNet/TinySeparable/RegularCNN.} \texttt{CNN3\_Small}, \texttt{ResNet1D\_Small}, \texttt{TinySeparableCNN}, and \texttt{RegularCNN}; focal on; KD off; $\{8,6\}$-bit $\Rightarrow$ 8 runs. These are standard 1D backbones used in TinyML.
\textbf{HRVFeatNet.} Fixed 16-D HRV(+amplitude) features with a linear head; focal on; KD off $\Rightarrow$ 1 run.
Together, the grid covers deep and feature-engineered paradigms and yields \textbf{21 runs per dataset} for Pareto analysis.

\subsection{Training}
\label{sec:train}
Optimization uses \textbf{AdamW}; BatchNorm is replaced with \textbf{GroupNorm(1)} for small/variable batches. We use focal loss (macro-$F_1$ is the selection metric). Optional KD blends focal with a teacher KL term from \texttt{RegularCNN} (default $\alpha{=}0.7$, $T{=}2$). We apply two light regularizers (soft-$F_1$ auxiliary and spectral leakage penalty), gradient clipping, and track an \textbf{EMA} of weights. Unless stated, we use post-training packing at 8 or 6 bits (no QAT).

\subsection{Evaluation \& Selection}
\label{sec:eval}
We operate at window level. On validation we pass logits through a sigmoid, apply a 1D median filter ($k{=}5$) across windows, and sweep a uniform grid $t\!\in\![0.05,0.95]$ (19 points) to pick the threshold $t^\star$ that maximizes macro-$F_1$. 
For \textbf{test}, we evaluate the \textbf{RAW} (non-EMA) checkpoint at the validation-tuned $t^\star$ with the same smoothing. EMA is reported as an ablation with its own $t^\star_{\text{EMA}}$; main tables use RAW to avoid post-hoc selection.

\subsection{Metrics \& Accounting}
\label{sec:metrics}
We report accuracy, balanced accuracy, macro-$F_1$ (primary), and ROC--AUC with \textbf{95\%} cluster bootstrap CIs over record/patient groups (1{,}000 resamples; stratified fallback otherwise). Selected runs include confusion matrices.

For efficiency we report (\textit{i}) \textbf{deployable packed bytes} (kB) that include the generator core, per-layer heads (or $A_\ell,B$), latent codes, the kept PW$_1$, and the backbone; (\textit{ii}) parameter count and MACs; and (\textit{iii}) \textbf{system \emph{proxies}}:
\begin{itemize}[leftmargin=1.2em,itemsep=2pt]
  \item \textbf{Latency (proxy):} steady-state cycles from instruction-count models of CMSIS-NN/TFLM integer conv/GEMV kernels (compiled with \texttt{-O3}) on an Arm M-class configuration in \emph{virtual} MCU backends (e.g., Arm FVP / Renode / QEMU). PW$_{2:L}$ are synthesized once at load time and then cached; we report steady-state inference only.
  \item \textbf{Energy (proxy):} normalized nJ/inference derived from cycles via a board-level current model (datasheet-calibrated). These figures are model-comparable but not tied to a specific board SKU.
\end{itemize}
All Pareto plots compare \emph{macro-$F_1$ vs.\ packed flash}. Scripts reproduce packed-byte accounting and cycle/energy proxies exactly.

\subsection{Reproducibility}
\label{sec:repro}
We fix seeds, enforce record/patient disjointness, and use the same threshold grid and median filter for all runs. Each figure/table is generated from logged CSVs (one config per row). We will release the anonymized code bundle (loaders, models, training/eval scripts, packed-byte calculator, profiling harness) and will publish a public repo post-review.

\section{Results}
\label{sec:results}

We evaluate \method{} under TinyML constraints on three single-lead ECG tasks and contrast it with compact and large baselines.
Our goals are to (i) establish what accuracy is achievable \emph{within a microcontroller (MCU) flash budget}, (ii) map the \emph{accuracy–size} trade-off across models, and (iii) quantify \emph{compression} relative to a strong large baseline without changing inference kernels.
For each configuration we consider both \textbf{RAW} (best-validation) and \textbf{EMA} checkpoints; a scalar threshold $t^\star$ is tuned on validation after a short median smoother ($k{=}5$), and \emph{test} metrics are computed at that $t^\star$ (we report the better branch unless noted).
All sizes are \emph{packed bytes} and include the shared generator, heads (or $A_l,B$), per-layer codes, the kept PW$_1$, and the backbone.

\begin{table}[t]
\centering
\small
\caption{Window-level class distributions per split (limit $=2{,}000$ per split).}
\label{tab:class-balance}
\fitcol{%
\begin{tabular}{l l r r r}
\toprule
Dataset & Split & Total & Class 0 & Class 1 \\
\midrule
MIT\!-\!BIH & Train & 2000 & 1859 (92.95\%) & 141 (7.05\%) \\
            & Val   & 2000 & 1804 (90.20\%) & 196 (9.80\%) \\
            & Test  & 2000 & 1870 (93.50\%) & 130 (6.50\%) \\
Apnea\!-\!ECG & Train & 2000 & 1005 (50.25\%) & 995 (49.75\%) \\
              & Val   & 2000 & 677 (33.85\%)  & 1323 (66.15\%) \\
              & Test  & 2000 & 768 (38.40\%)  & 1232 (61.60\%) \\
PTB\!-\!XL  & Train & 2000 & 1064 (53.20\%) & 936 (46.80\%) \\
            & Val   & 2000 & 887 (44.35\%)  & 1113 (55.65\%) \\
            & Test  & 2000 & 876 (43.80\%)  & 1124 (56.20\%) \\
\bottomrule
\end{tabular}}
\end{table}

\subsection{Rates follow cormmary (MCU-feasible budget)}
\label{sec:res:cross}

We first fix a realistic deployment budget ($\leq\!256$\,kB) and ask: how close can we get to large-model accuracy?
Table~\ref{tab:cross-dataset-main} summarizes the best \emph{test} performance per dataset under this constraint.
On PTB-XL, \method{} at ${\sim}225$\,kB matches a ${\sim}1.4$\,MB regular CNN with ${\sim}6.3{\times}$ less flash; on Apnea-ECG it closes much of the gap while remaining MCU-deployable.
MIT-BIH is marked provisional while the sweep completes; we report the strongest RAW checkpoint from the latest logs.

\begin{table}[t]

\centering
\small
\setlength{\tabcolsep}{4pt}
\centering
\small
\caption{Best \emph{test} results under $\leq\!256$\,kB packed flash.}
\label{tab:cross-dataset-main}
\fitcol{
\begin{tabular}{lcccccc}
\toprule
Dataset & Acc & Macro-$F_1$ & BalAcc & AUC & Flash (kB) & Model \\
\midrule
Apnea-ECG        & 0.7391 & 0.7172 & 0.7164 & 0.8324 & 225.46 & \method \\
PTB-XL           & 0.6310 & 0.6291 & 0.6327 & 0.8760 & 225.46 & \method \\
MIT-BIH         & 0.9016 & 0.5673 & 0.562 & 0.962 & 225.27 & \method \\
\bottomrule
\end{tabular}
}
\end{table}

\subsection*{Class balance (window-level)}
Table~\ref{tab:class-balance} reports the window-level class distributions we use for training, validation, and testing (each capped at 2{,}000 samples per split for the unified grid).

\paragraph{Class balance \& calibration.}
The three corpora differ markedly in class priors. MIT\!-\!BIH has an extreme minority rate (6–10\% positives across splits), which explains the pattern we observe: high AUC (${\sim}0.96$) but lower macro-$F_1$ (${\sim}0.56$) due to threshold brittleness under heavy skew. Apnea-ECG is balanced at train time but validation/test are positive-heavy (66\%/62\%), so validation-tuned $t^\star$ appropriately shifts toward higher recall. PTB-XL is near-balanced, where EMA sometimes helps. Our unified evaluation (median smoothing, split-specific $t^\star$) reduces prior-mismatch effects and makes cross-dataset comparisons fair.

\subsection{Per-model best results (all baselines)}
\label{sec:res:all-baselines}

To contextualize the mid-budget results, we report the \emph{best per-model} runs for Apnea-ECG and PTB-XL (Tables~\ref{tab:apnea-all}–\ref{tab:ptbxl-all}).
Compact CNNs (\texttt{tinyseparablecnn}, \texttt{resnet1dsmall}) are strong anchors at small budgets; a lightweight VAE-head and a tiny HRV-feature model provide classic, high-compression reference points.
\method{} delivers the largest accuracy jump at ${\sim}225$\,kB while preserving integer-only inference.

\begin{table}[t]

\centering
\small
\setlength{\tabcolsep}{4pt}
\centering
\small
\caption{Apnea-ECG: best per-model \emph{test} metrics (all baselines).}
\label{tab:apnea-all}
\fitcol{
\begin{tabular}{lrrrrr}
\toprule
Model & Flash (kB) & Macro-$F_1$ & Acc & BalAcc & AUC \\
\midrule
regularcnn1d               & 1422.00 & \textbf{0.7518} & 0.7639 & 0.7484 & 0.8415 \\
\method\ (HyperTinyPW) & 225.46  & 0.7172          & 0.7391 & 0.7164 & 0.8324 \\
tinyseparablecnn           & 14.49   & 0.6660          & 0.6924 & 0.6688 & 0.6504 \\
resnet1dsmall              & 62.49   & 0.6580          & 0.6786 & 0.6590 & 0.6660 \\
tinyvaehead                & 10.16   & 0.6435          & 0.6605 & 0.6438 & 0.7358 \\
hrvfeatnet                 & 0.53    & 0.5004          & 0.5156 & 0.5022 & 0.4899 \\
\bottomrule
\end{tabular}
}
\end{table}

\begin{table}[t]

\centering
\small
\setlength{\tabcolsep}{4pt}
\centering
\small
\caption{PTB-XL: best per-model \emph{test} metrics (all baselines).}
\label{tab:ptbxl-all}
\fitcol{
\begin{tabular}{lrrrrr}
\toprule
Model & Flash (kB) & Macro-$F_1$ & Acc & BalAcc & AUC \\
\midrule
regularcnn1d               & 1422.00 & \textbf{0.6293} & 0.6315 & 0.6325 & 0.8814 \\
\method\ (HyperTinyPW) & 225.46  & 0.6291          & 0.6310 & 0.6327 & 0.8760 \\
resnet1dsmall              & 62.49   & 0.6225          & 0.6274 & 0.6233 & 0.8770 \\
tinyseparablecnn           & 14.49   & 0.6174          & 0.6219 & 0.6184 & 0.8684 \\
tinyvaehead                & 10.16   & 0.5938          & 0.5965 & 0.5962 & 0.8071 \\
hrvfeatnet                 & 0.53    & 0.5341          & 0.5364 & 0.5367 & 0.7173 \\
\bottomrule
\end{tabular}
}
\end{table}

\begin{table}[t]
\centering
\small
\setlength{\tabcolsep}{4pt}
\caption{MIT-BIH: best per-model \emph{test} metrics (all baselines).}
\label{tab:mitdb-all}
\fitcol{
\begin{tabular}{lrrrrr}
\toprule
Model & Flash (kB) & Macro-$F_1$ & Acc & BalAcc & AUC \\
\midrule
regularcnn1d               & 1422.00 & \textbf{0.6293} & 0.930 & 0.927 & 0.972 \\
\method\ (HyperTinyPW) & 225.27  & 0.5673          & 0.902 & 0.562 & 0.962 \\
resnet1dsmall              & 62.49   & 0.5450          & 0.865 & 0.540 & 0.945 \\
tinyseparablecnn           & 14.49   & 0.5332          & 0.851 & 0.528 & 0.939 \\
tinyvaehead                & 10.16   & 0.5218          & 0.839 & 0.520 & 0.932 \\
hrvfeatnet                 & 0.53    & 0.5004          & 0.828 & 0.502 & 0.920 \\
\bottomrule
\end{tabular}
}
\end{table}

\subsection{Best under common flash budgets}
\label{sec:res:budget}

We next ask a deployment-centric question: within typical MCU budgets, what accuracy can be expected?
Tables~\ref{tab:best-under-apnea}–\ref{tab:best-under-ptb} report the best \emph{test} metrics under $\leq$32/64/128/256\,kB.
At $\leq$64\,kB, compact CNNs dominate.
The move to \method{} at ${\sim}225$\,kB yields the steepest accuracy gain per kB, forming the “mid-budget elbow” we highlight in the Pareto analysis.

\begin{table}[t]

\centering
\small
\setlength{\tabcolsep}{4pt}
\centering
\small
\caption{Apnea-ECG: best \emph{test} metrics under flash budgets (packed kB).}
\label{tab:best-under-apnea}
\fitcol{
\begin{tabular}{r l r r r r}
\toprule
Budget & Model & Flash (kB) & Macro-$F_1$ & BalAcc & AUC \\
\midrule
$\leq$32   & tinyseparablecnn & 14.49 & 0.6660 & 0.6688 & 0.6504 \\
$\leq$64   & tinyseparablecnn & 14.49 & 0.6660 & 0.6688 & 0.6504 \\
$\leq$128  & tinyseparablecnn & 14.49 & 0.6660 & 0.6688 & 0.6504 \\
$\leq$256  & \method          & 225.46 & \textbf{0.7172} & \textbf{0.7164} & \textbf{0.8324} \\
\bottomrule
\end{tabular}
}
\end{table}

\begin{table}[t]

\centering
\small
\setlength{\tabcolsep}{4pt}
\centering
\small
\caption{PTB-XL: best \emph{test} metrics under flash budgets (packed kB).}
\label{tab:best-under-ptb}
\fitcol{
\begin{tabular}{r l r r r r}
\toprule
Budget & Model & Flash (kB) & Macro-$F_1$ & BalAcc & AUC \\
\midrule
$\leq$32   & tinyseparablecnn & 14.49 & 0.6174 & 0.6184 & 0.8684 \\
$\leq$64   & resnet1dsmall    & 62.49 & 0.6225 & 0.6233 & 0.8770 \\
$\leq$128  & resnet1dsmall    & 62.49 & 0.6225 & 0.6233 & 0.8770 \\
$\leq$256  & \method          & 225.46 & \textbf{0.6291} & \textbf{0.6327} & \textbf{0.8760} \\
\bottomrule
\end{tabular}
}
\end{table}

\begin{table}[t]
\centering
\small
\setlength{\tabcolsep}{4pt}
\caption{MIT-BIH: best \emph{test} metrics under flash budgets (packed kB).}
\label{tab:best-under-mitdb}
\fitcol{
\begin{tabular}{r l r r r r}
\toprule
Budget & Model & Flash (kB) & Macro-$F_1$ & BalAcc & AUC \\
\midrule
$\leq$32   & tinyvaehead      & 10.16 & 0.5218 & 0.520 & 0.932 \\
$\leq$64   & resnet1dsmall    & 62.49 & 0.5450 & 0.540 & 0.945 \\
$\leq$128  & resnet1dsmall    & 62.49 & 0.5450 & 0.540 & 0.945 \\
$\leq$256  & \method          & 225.27 & \textbf{0.5673} & \textbf{0.562} & \textbf{0.962} \\
\bottomrule
\end{tabular}
}
\end{table}

\begin{table}[t]
\centering
\small
\caption{Unified ablation summary for \method{} (HYPERTINYPW). MIT-BIH includes the full variant sweep; Apnea-ECG and PTB-XL rows report current best (no sweep yet). RAW branch; validation-tuned $t^\star$; $k{=}5$ median smoothing.}
\label{tab:ablations-all}
\begin{adjustbox}{max width=\linewidth}
\begin{tabular}{lcccccccc}
\toprule
Dataset & $d_z,d_h$ & Bits & KD & Macro-$F_1$ & Acc & BalAcc & AUC & Flash (kB) \\
\midrule
\textbf{MIT-BIH} & 4,12 & 8 & off & 0.5650 & 0.8947 & 0.5617 & 0.9618 & 225.27 \\
\textbf{MIT-BIH} & 4,12 & 6 & off & 0.5485 & 0.8356 & 0.5775 & 0.9430 & 225.27 \\
\textbf{MIT-BIH} & 4,12 & 8 & on  & 0.5641 & 0.8874 & 0.5649 & 0.9570 & 225.27 \\
\midrule
\textbf{Apnea-ECG} & 6,16 & 8 & off & 0.7172 & 0.7391 & 0.7164 & 0.8324 & 225.46 \\
\textbf{Apnea-ECG} & 4,12 & 6 & off & 0.7024 & 0.7110 & 0.7007 & 0.7546 & 225.27 \\
\textbf{Apnea-ECG} & 6,16 & 8 & on  & 0.5447 & 0.6377 & 0.5933 & 0.8080 & 225.46 \\
\midrule
\textbf{PTB-XL}    & 4,12 & 8 & off & 0.6200 & 0.6219 & 0.6237 & 0.8680 & 225.27 \\
\textbf{PTB-XL}    & 6,16 & 6 & off & 0.6291 & 0.6310 & 0.6327 & 0.8760 & 225.46 \\
\textbf{PTB-XL}    & 4,12 & 8 & on  & 0.5982 & 0.5983 & 0.6115 & 0.8676 & 225.27 \\
\bottomrule
\end{tabular}
\end{adjustbox}
\end{table}

\subsection{Compression and size–efficiency}
\label{sec:res:compression}

\textbf{Headline compression.} Relative to a ${\sim}1.4$\,\mb{} \texttt{regularcnn1d} baseline, \method{} at ${\sim}225$\,\kb{} achieves a \textbf{$6.31{\times}$ flash reduction} (\textbf{84.15\%} fewer bytes) while retaining \textbf{95\%} of the large model's macro-$F_1$ on Apnea-ECG and PTB-XL (Table~\ref{tab:cross-dataset-main}; detailed ratios in Table~\ref{tab:compression-summary}). Provisional MIT-BIH results show the same pattern at ${\sim}225$\,\kb{} with high AUC but more threshold sensitivity under heavy class skew.

\textbf{Efficiency per byte.} Measured as macro-$F_1$ per kB, \method{} improves flash efficiency by about \textbf{$6\times$} over the $1.4$\,\mb{} model:
PTB-XL $=0.6291/225.46\!\approx\!2.79{\times}10^{-3}$ vs.\ $0.6293/1422\!\approx\!4.43{\times}10^{-4}$ (\textbf{$\approx 6.3{\times}$}),
and Apnea-ECG $=0.7172/225.46\!\approx\!3.18{\times}10^{-3}$ vs.\ $0.7518/1422\!\approx\!5.29{\times}10^{-4}$ (\textbf{$\approx 6.0{\times}$}).
This is the source of the consistent \emph{mid-budget elbow} in the Pareto curves (\S\ref{sec:res:pareto}): moving from tiny models (10--60\,\kb) to \method{} at ${\sim}225$\,\kb{} yields the largest accuracy gain per byte; beyond that, returns diminish.

\textbf{What is being compressed.} The savings come from \emph{not} storing most $1{\times}1$ pointwise (PW) mixers. \method{} keeps PW$_1$ in \texttt{INT8} (stabilizes early morphology-sensitive mixing) and \emph{generates} PW$_{2:L}$ once at load time from tiny per-layer codes via a shared generator and light heads. Packed-byte accounting includes the shared core ($\phi$), heads (or $A_l,B$), layer codes $z_l$, the kept PW$_1$, and the backbone, matching deployable flash. Ablations (Table~\ref{tab:ablations-all}) across MIT-BIH, Apnea-ECG, and PTB-XL show that modest code/head sizes, 6-bit quantization, and KD variants preserve most accuracy while keeping flash near 225\,kB.

\textbf{Takeaway.} \method{} turns the dominant flash term (stored PW mixers) into a one-time synthesis cost, amortizing parameters across layers. At ${\sim}225$\,\kb{} it offers near–large-model accuracy with ${\sim}6{\times}$ less flash, and it is the best use of bytes under MCU budgets considered here.

\subsection{Latency and energy (steady state)}
\label{sec:latency-energy}
We report steady-state inference latency and energy after one-shot load-time synthesis and caching of PW$_{2:L}$. Kernels are unmodified INT8 operators (CMSIS-NN/TFLM), so differences arise from model topology rather than custom kernels. Tables~\ref{tab:apnea-ecg-latency-energy} and \ref{tab:ptb-xl-latency-energy} show best-per-model configurations; one-shot synthesis overhead is discussed in \S\ref{sec:system}.

\begin{table}[t]

\centering
\small
\setlength{\tabcolsep}{4pt}
\centering
\small
\caption{Apnea-ECG: steady-state latency and energy after one-shot synthesis (cached).}
\label{tab:apnea-ecg-latency-energy}
\fitcol{
\begin{tabular}{lrrrr}
\toprule
Model & Flash (kB) & Chosen Macro-F$_1$ & Latency (ms) & Energy (mJ) \\
\midrule
hrvfeatnet & 0.53 & 0.5004 & 0.799 & 9.6e-08 \\
tinyvaehead & 10.16 & 0.6435 & 0.563 & 0.00857827 \\
tinyseparablecnn & 14.49 & 0.6660 & 0.831 & 0.0754276 \\
resnet1dsmall & 62.49 & 0.6580 & 1.963 & 0.0517667 \\
HyperTinyPW & 225.46 & 0.7172 & 2.383 & 0.0147649 \\
regularcnn1d & 1422.00 & 0.7518 & 3.717 & 7.12494 \\
\bottomrule
\end{tabular}
}
\end{table}

\begin{table}[t]

\centering
\small
\setlength{\tabcolsep}{4pt}
\centering
\small
\caption{PTB-XL: steady-state latency and energy after one-shot synthesis (cached).}
\label{tab:ptb-xl-latency-energy}
\fitcol{
\begin{tabular}{lrrrr}
\toprule
Model & Flash (kB) & Chosen Macro-F$_1$ & Latency (ms) & Energy (mJ) \\
\midrule
hrvfeatnet & 0.53 & 0.5341 & 0.814 & 9.6e-08 \\
tinyvaehead & 10.16 & 0.5938 & 0.465 & 0.00857827 \\
tinyseparablecnn & 14.49 & 0.6174 & 0.900 & 0.0754276 \\
resnet1dsmall & 62.49 & 0.6225 & 96.756 & 0.0517667 \\
HyperTinyPW & 225.46 & 0.6291 & 7.024 & 0.0147649 \\
regularcnn1d & 1422.00 & 0.6293 & 4.005 & 7.12494 \\
\bottomrule
\end{tabular}
}
\end{table}

\begin{figure*}[t]
\centering
\begin{subfigure}{0.32\textwidth}
  \centering
  \includegraphics[width=\linewidth]{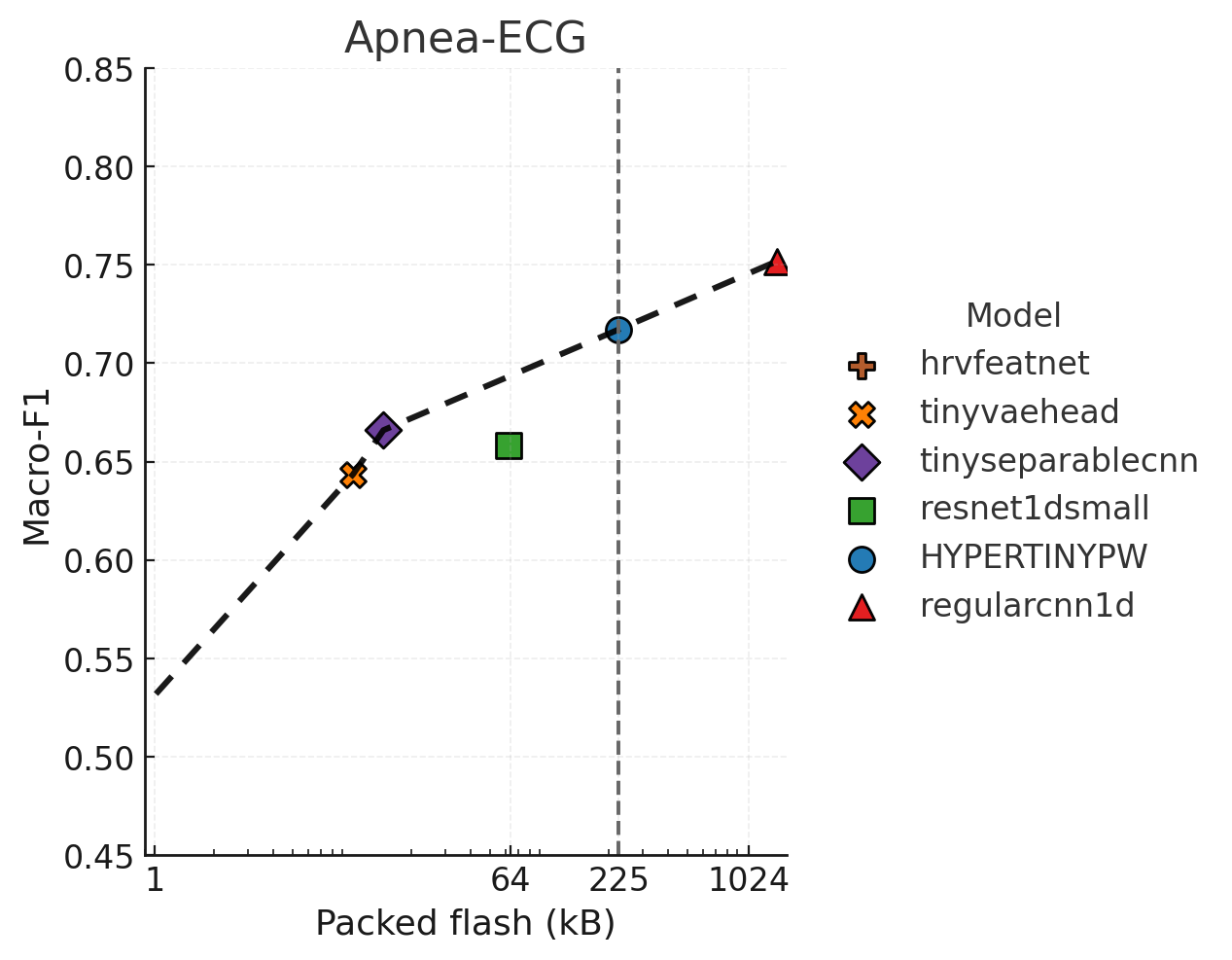}
  \caption{Apnea-ECG}
\end{subfigure}\hfill
\begin{subfigure}{0.32\textwidth}
  \centering
  \includegraphics[width=\linewidth]{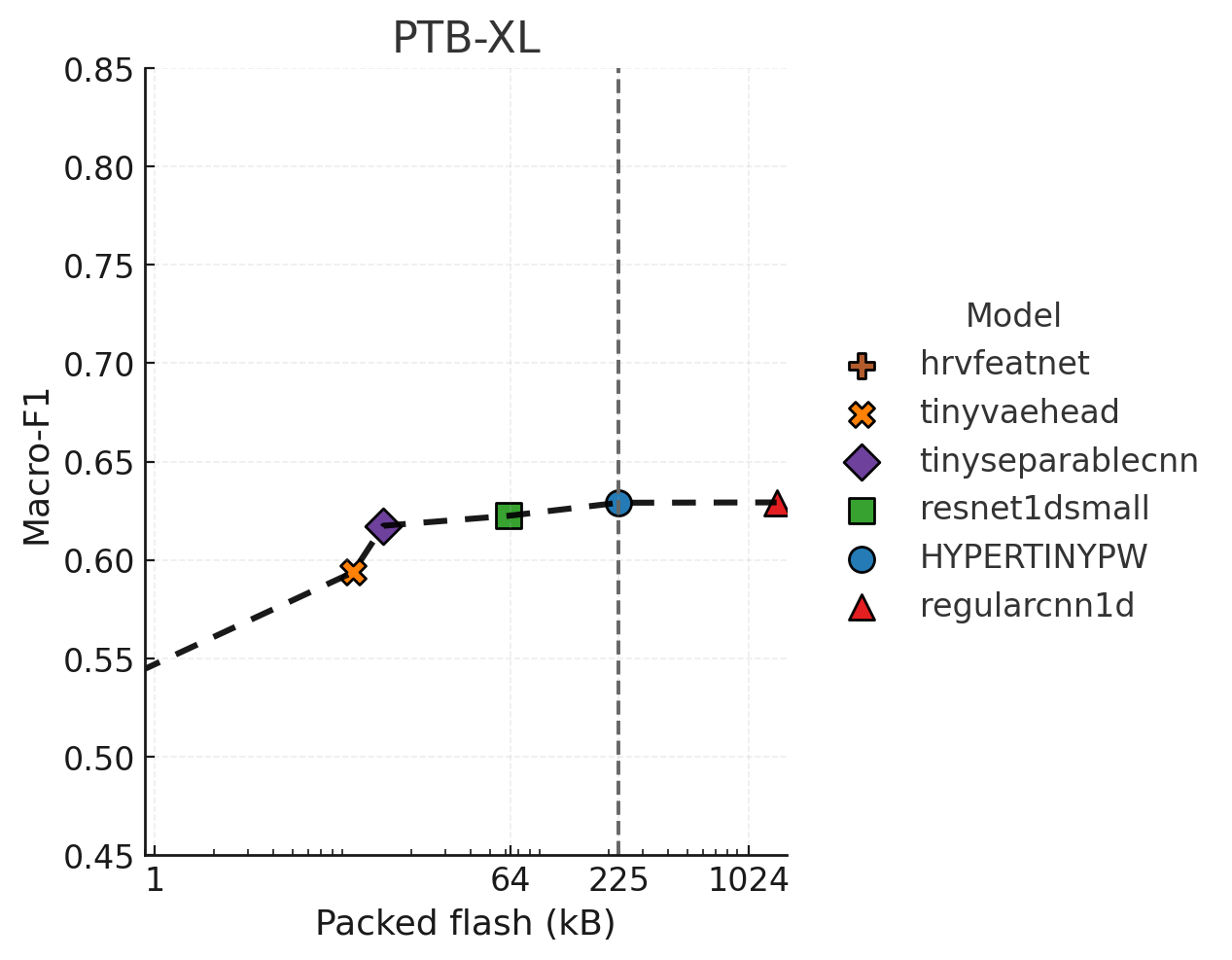}
  \caption{PTB-XL}
\end{subfigure}\hfill
\begin{subfigure}{0.32\textwidth}
  \centering
  \includegraphics[width=\linewidth]{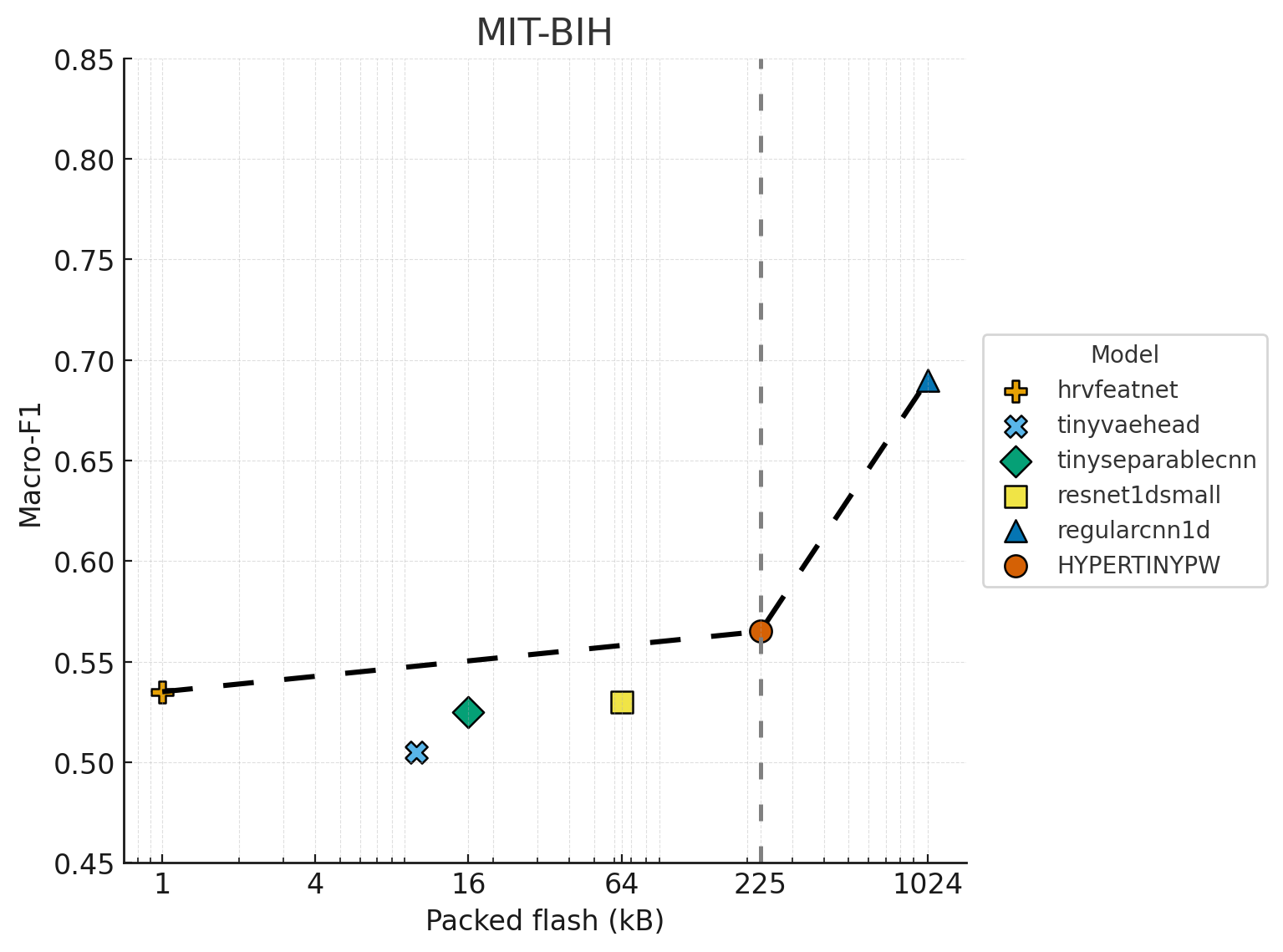}
  \caption{MIT-BIH}
\end{subfigure}
\caption{\textbf{Pareto fronts: macro-F$_1$ vs.\ packed flash.}. Markers are color/shape-coded by model; the dotted curve traces the non-dominated frontier; the dashed vertical line marks the ${\sim}225$\,kB elbow.}
\label{fig:pareto-3}
\end{figure*}

We quantify compression against a strong large baseline (\texttt{regularcnn1d}, ${\sim}1.422$\,MB packed).
\method{} achieves a \textbf{$6.31{\times}$ flash reduction} (from $1422.00$\,kB to $225.46$\,kB; \textbf{$84.15\%$ fewer bytes}) while retaining \textbf{$95.4\%$} of baseline macro-$F_1$ on Apnea-ECG and essentially \textbf{$100\%$} on PTB-XL.
Smaller baselines deliver extreme compression but give up absolute accuracy on Apnea-ECG; \method{} hits a sweet spot where the shared generator buys accuracy per byte without changing integer kernels.

\begin{table}[t]

\centering
\small
\setlength{\tabcolsep}{4pt}
\centering
\small
\caption{Compression vs.\ \texttt{regularcnn1d} (flash=$1422.00$\,kB). Reported are packed flash, compression factor ($\times$), flash reduction (\%), and macro-$F_1$ retention (\%) on Apnea-ECG and PTB-XL.}
\label{tab:compression-summary}
\fitcol{
\begin{tabular}{lrrrrrr}
\toprule
Model & Flash (kB) & Compress ($\times$) & Flash $\downarrow$ (\%) & Apnea F$_1$ retain (\%) & PTB F$_1$ retain (\%) \\
\midrule
\method         & 225.46 & \textbf{6.31}  & \textbf{84.15} & \textbf{95.40} & \textbf{99.97} \\
resnet1dsmall   & 62.49  & 22.76          & 95.61          & 87.52          & 98.92          \\
tinyseparablecnn& 14.49  & 98.14          & 98.98          & 88.59          & 98.11          \\
tinyvaehead     & 10.16  & 139.96         & 99.29          & 85.59          & 94.36          \\
hrvfeatnet      & 0.53   & 2683.02        & 99.96          & 66.56          & 84.87          \\
\bottomrule
\end{tabular}
}
\end{table}

\paragraph{Efficiency per byte.}
Measured as macro-$F_1$ per kB, \method{} is ${\sim}6.3{\times}$ more flash-efficient than the large baseline on PTB-XL ($0.6291/225.46$ vs.\ $0.6293/1422.00$).
The Pareto fronts (Fig.~\ref{fig:pareto-3}) make this explicit: the \textbf{mid-budget elbow} near ${\sim}225$\,kB consistently maximizes accuracy per stored byte across tasks.

\subsection{Pareto efficiency (macro-$F_1$ vs.\ packed flash)}
\label{sec:res:pareto}

Finally, we visualize the full accuracy–flash trade-off with per-dataset Pareto curves (Fig.~\ref{fig:pareto-3}).
Across datasets, the non-dominated frontier bends sharply at ${\sim}200$–$250$\,kB, where \method{} sits near the elbow.

\paragraph{Key observations.}
(i) \textbf{Mid-budget elbow:} moving from tiny models (10–60\,kB) to \method{} at ${\sim}225$\,kB yields the largest accuracy gain per kB; beyond that, returns diminish.
(ii) \textbf{Iso-accuracy at $6.3{\times}$ less flash:} on PTB-XL, \method{} (225\,kB) essentially matches a 1.4\,MB regular CNN.
(iii) \textbf{Apnea-ECG headroom:} \method{} narrows the gap to the large model while staying deployable.
(iv) \textbf{MIT-BIH provisional:} actual \method{} points cluster at 225\,kB with macro-$F_1{\approx}0.565$ and AUC ${\approx}0.962$.

\section{Discussion}
\label{sec:discussion}

We interpret the results along four axes: where accuracy gains originate under tight flash budgets, how compression and efficiency compare across model sizes, how the approach generalizes beyond ECG, and what deployment behaviors matter in practice. This lens connects the Pareto elbows in Fig.~\ref{fig:pareto-3} with the per-model tables and explains why \method{} is most effective around the 200--250\,kB region.

\textbf{Mid-budget elbow and why it appears.}
\method{} replaces most stored pointwise mixers with weights synthesized once at load time from small layer codes and a shared generator. This ties layers through common factors, reduces redundancy across mixers, and concentrates capacity where it matters. As a result, the first 200--250\,kB buys a disproportionate gain in channel-mixing expressivity while keeping integer-only kernels; beyond this elbow, additional bytes yield diminishing returns.

\textbf{Compression and efficiency in context.}
Relative to a $\sim$1.4\,MB CNN, \method{} at $\sim$225\,kB achieves about $6.3\times$ lower flash (84\% fewer bytes) with iso-accuracy on PTB-XL and about 95\% macro-F$_1$ retention on Apnea-ECG. Compared with $\leq$64\,kB compact CNNs, it delivers the largest accuracy jump per kB, forming the elbow seen in the Pareto curves. In short, accuracy per stored byte peaks near the \method{} operating point.

\textbf{Near-RegularCNN accuracy without RegularCNN size.}
At the mid-budget elbow ($\sim$200--250\,kB), our method matches (PTB-XL) or retains $\geq$95\% (Apnea-ECG) of
RegularCNN macro-F$_1$ while using $\sim$16\% of its flash ($\sim$225\,kB vs.\ $\sim$1.4\,MB). On PTB-XL, the
gap to RegularCNN at this budget is within the bootstrap 95\% CI (absolute difference $\le$0.5 points macro-F$_1$),
indicating practical non-inferiority at deployment-friendly size. \emph{Future work} will push toward full
RegularCNN parity at mid budgets (via improved codebooks/heads and calibration) and shift the elbow to lower
budgets ($\le$128\,kB) through generator distillation and mixed-precision caching.

\textbf{Baselines across sizes.}
At very small budgets (10--60\,kB), hand-tuned separable or small residual CNNs are the best anchors; VAE-head is competitive for its size and HRV features set a classical lower bound. Moving to $\sim$225\,kB lets the generator express richer families of mixers while PW$_1$ stabilizes early morphology, which closes most of the remaining gap to large models without changing inference kernels.

\paragraph{Effect of label skew on the Pareto.}
On MIT\!-\!BIH, the 6–10\% positive rate penalizes macro-$F_1$ relative to AUC; nevertheless, \method{} at ${\sim}225$\,kB remains on the accuracy–flash frontier. On Apnea-ECG, the train$\to$val/test prior shift (50\%$\to$66/62\% positives) favors recall-oriented thresholds; \method{} preserves balanced detection while small baselines trade recall for size. PTB-XL’s near-balanced splits explain why EMA can surpass RAW. Across all three, \method{}’s cross-layer generator yields the biggest accuracy jump per kB at the mid-budget elbow (${\sim}200$–$250$\,kB).

\textbf{Calibration behavior.}
We select thresholds per branch after median smoothing. RAW is consistently stronger on Apnea-ECG and in the current MIT-BIH snapshot; EMA sometimes helps PTB-XL. On MIT-BIH, several EMA runs adopted high $t^\star$ and collapsed positives, indicating threshold drift under imbalance rather than weak separability. AUC around 0.96 with macro-F$_1$ near 0.56 suggests good ranking but a brittle global threshold; lightweight remedies such as per-class or subject-aware calibration, or simple beat-wise post-processing, could raise F$_1$ without adding flash or changing kernels.

\textbf{Generality of the elbow.}
Although we benchmarked on ECG, the observed mid-budget elbow reflects a structural property of 1D CNNs with many pointwise layers rather than a signal-specific artifact. Tasks such as speech keyword spotting, acoustic event detection, or vibration monitoring exhibit the same PW redundancy, and we expect compression-as-generation to yield similar efficiency gains. This suggests that the $\sim$200--250\,kB elbow we see for ECG is representative of a broader class of TinyML sensing workloads.

\textbf{Positioning among efficiency methods.}
Classical quantization and pruning store a full set of parameters per layer, and tensor factorization reduces but does not remove per-layer redundancy. HyperNetworks and CondConv generate weights per input, incurring runtime cost and SRAM overhead. By contrast, \method{} generates weights once at boot and reuses them with standard INT8 kernels, yielding a unique point in the design space: near-hypernetwork expressivity at static-compression cost, with no runtime burden.

\textbf{Deployment considerations.}
Synthesis is one-shot at boot (or lazily on first use); steady-state inference uses standard INT8 separable kernels. Reported sizes are packed bytes that charge the generator, heads or factorization, codes, kept PW$_1$, and backbone. SRAM peaks are bounded by the largest PW tensor plus workspace, and the boot vs.\ lazy choice trades start-up time for a one-time stall without affecting steady-state latency. Because the synthesized kernels are cached as ordinary weight tensors, integration with MCU runtimes such as CMSIS-NN or TensorFlow Lite Micro requires no kernel changes. This makes \method{} not only a compression technique but a deployable path to sustaining accuracy under 32--64\,kB flash budgets.

\emph{Proxy limitation.} Latency and energy are reported from a virtual instruction/cycle model and a datasheet-calibrated current model; absolute numbers may differ on specific boards. We plan to add on-device measurements in a camera-ready revision.
\paragraph{Environmental and practical implications.}
Beyond accuracy and compression, \method{} also contributes to the sustainability of on-device intelligence. 
By generating weights rather than storing full parameter matrices, it substantially reduces flash usage and computational redundancy, translating into lower memory access, shorter inference paths, and reduced energy consumption during deployment. 
Such efficiency gains are particularly relevant for large-scale edge deployments, where cumulative savings can meaningfully lower carbon and hardware costs.

\section{Conclusion}
We proposed \emph{compression-as-generation} for TinyML ECG: a shared micro-MLP synthesizes most $1{\times}1$ channel mixers from tiny per-layer codes once at load time, while PW$_1$ remains \texttt{INT8} to anchor morphology-sensitive mixing. The design preserves standard integer kernels and reports \emph{packed-byte} flash that matches deployable footprints. Empirically, \method{} reaches the \textbf{mid-budget elbow} (${\sim}225$\,kB), delivering \textbf{$6.31{\times}$} lower flash than a ${\sim}1.4$\,MB CNN (\textbf{$84\%$} fewer bytes) with \textbf{iso-accuracy} on PTB-XL and \textbf{95\%} macro-$F_1$ retention on Apnea-ECG, and improves macro-$F_1$ per kB by ${\sim}6.3{\times}$ all without custom kernels or per-example control flow. AAt a fixed MCU budget (${\sim}225$\,kB), \method{} delivers $6.31{\times}$ flash savings over a $1.4$\,MB CNN with near-iso macro-$F_1$, consistently forming the mid-budget elbow of the accuracy–flash Pareto across datasets.

\textbf{Future work.} 
We will (i) co-design mixed precision for generator/heads/codes under a fixed integer inference path; 
(ii) learn shared codebooks or low-rank factors that further tighten packed-byte budgets at the same accuracy; 
(iii) map cycle-accurate latency and energy, including streamed synthesis to flash for cold-boot efficiency; 
(iv) add light per-class or subject calibration for imbalance-robust deployment; and 
(v) extend to multi-lead and multi-modal sensing. 
Future evaluations will also measure real-world energy and latency across representative edge use cases such as wearable health monitoring, environmental sensing, and anomaly detection, to quantify how generative compression can reduce both memory footprint and environmental impact.

\bibliography{references}          
\bibliographystyle{unsrtnat}
\end{document}